\pdfoutput=1
\documentclass[11pt]{article}

\usepackage{emnlp2021}

\usepackage{times}
\usepackage{latexsym}

\usepackage{colortbl}

\newcommand\graycell{\cellcolor[rgb]{0.9,0.9,0.9}}

\usepackage[T1]{fontenc}
\usepackage[utf8]{inputenc}

\usepackage{microtype}

\usepackage{graphicx, amsmath, subcaption,amssymb, comment, bm, amsthm, bbm, xcolor, soul, capt-of, booktabs, svg}

\usepackage{multirow}

\graphicspath{{../images/}} %Setting the graphicspath

\title{Calibrate your listeners! \\
Robust communication-based training for pragmatic speakers}

\author{Rose E. Wang{\normalfont \textsuperscript{1}},$\;$ Julia White {\normalfont \textsuperscript{2}},$\;$ Jesse Mu{\normalfont \textsuperscript{1}},$\;$ Noah D.\ Goodman{\normalfont \textsuperscript{1,3}} \\
  Departments of \textsuperscript{1}Computer Science, \textsuperscript{2}Electrical Engineering and \textsuperscript{3}Psychology \\
  Stanford University \\
  {\tt \{rewang, jiwhite, muj, ngoodman\}@stanford.edu} \\
}

\begin{document}
\maketitle

\begin{abstract}
To be good conversational partners, natural language processing (NLP) systems should be trained to produce contextually useful utterances. 
Prior work has investigated training NLP systems with communication-based objectives, where a neural listener stands in as a communication partner. 
However, these systems commonly suffer from semantic drift where the learned language diverges radically from natural language. 
We propose a method that uses a population of neural listeners to regularize speaker training.
We first show that language drift originates from  the poor uncertainty calibration of a neural listener, which makes high-certainty predictions on novel sentences. 
We explore ensemble- and dropout-based populations of listeners and find that the former results in better uncertainty quantification. 
We evaluate both population-based objectives on reference games, and show that the ensemble method with better calibration enables the speaker to generate pragmatic utterances while scaling to a large vocabulary and generalizing to new games and listeners.\footnote{The accompanying code can be found here: \hyperlink{https://github.com/rosewang2008/calibrate_your_listeners}{https://github.com/rosewang2008/calibrate\_your\_listeners}.}

\end{abstract}

\section{Introduction}
\begin{figure}[t]
    \centering
    \newcommand{\gw}{70mm}
    \newcommand{\plotr}{0.95}
    \begin{subfigure}[b]{\plotr\columnwidth}
      \centering
        % \vspace{-2em}
        % \hspace{-/2em}
        \includegraphics[width=\gw]{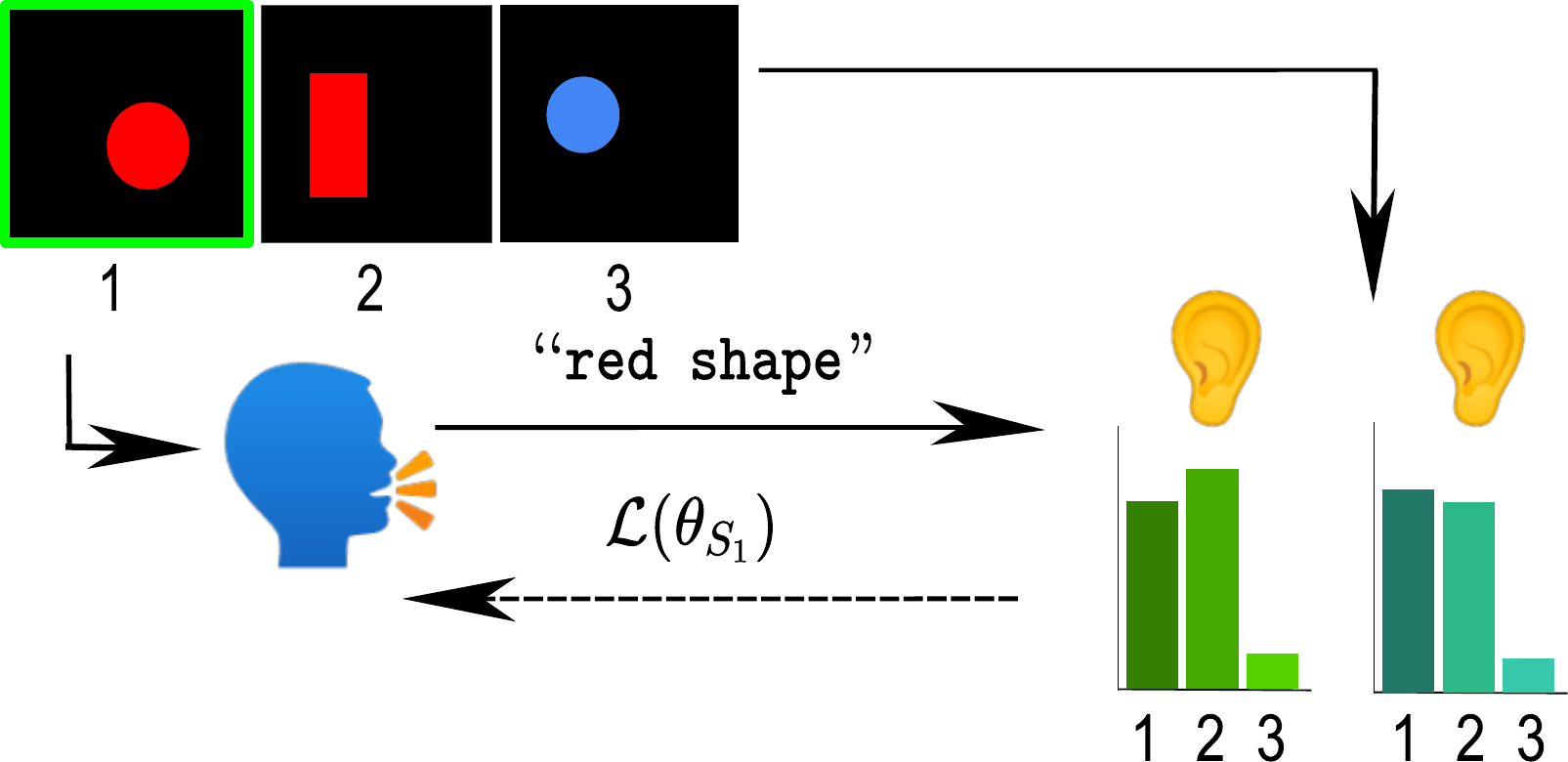}
        \caption{Communication-based objectives, where a speaker is trained with a listener reward model.}
    \end{subfigure}
    \begin{subfigure}[b]{\plotr\columnwidth}
      \centering
    \vspace{1em}
        \hspace{-2em}
        \includegraphics[width=65mm]{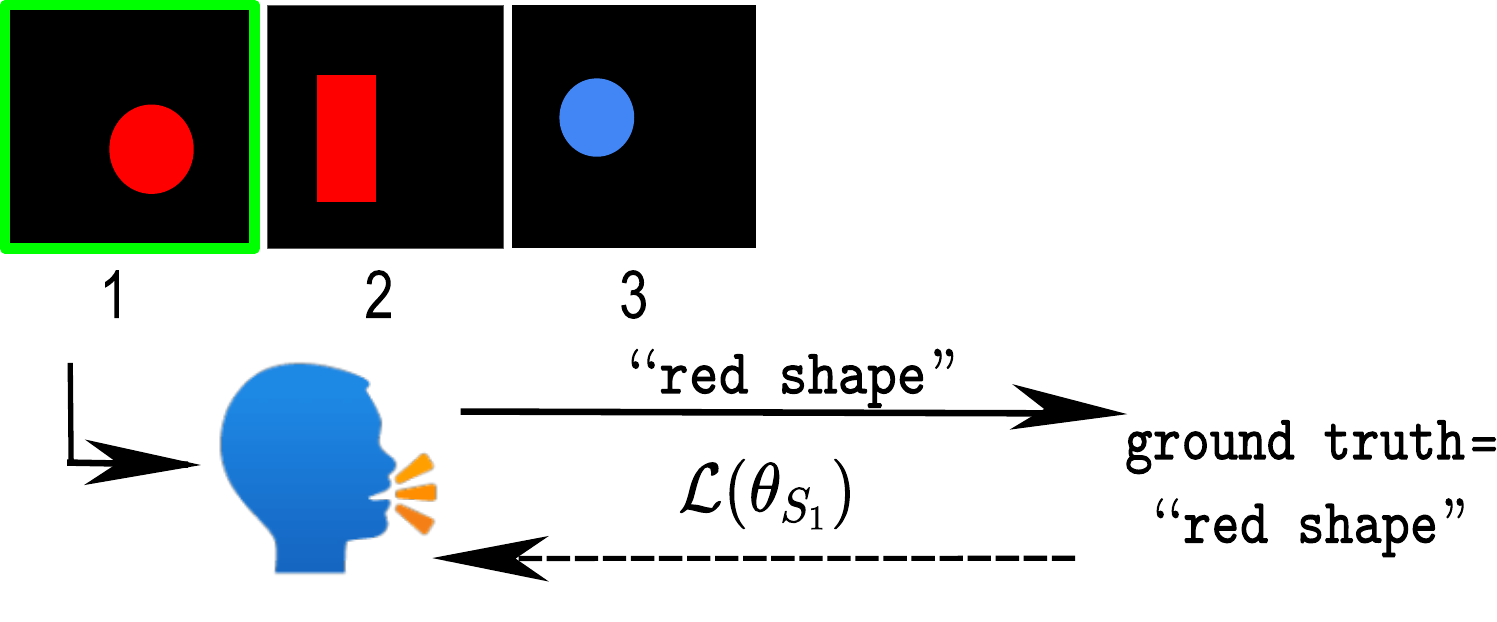}
        % \vspace{-5em}
        \caption{Traditional LM objectives, where a speaker is trained from a corpus of image descriptions.}
    \end{subfigure}
    \caption{The goal of the speaker (blue agent) is to generate context-aware utterances to refer to the image 1 and disambiguate from images 2 \& 3. Our work uses (a) communication-based training objectives rather than (b) non communication-based objectives which rely on a corpus of ground-truth captions.   \label{fig:differences}}
    \vspace{-0.5em}
\end{figure}

To be good conversational partners, language models (LMs) should learn to produce fluent utterances that serve the needs of their listeners \cite{Grice1975}.
However, they are often trained to capture the statistical, rather than communicative, properties of language via supervised learning \cite{bengio2003neural, radford2019language}.
Consider the reference game \cite{Lewis1969} in Figure~\ref{fig:differences}, where the goal of the speaker is to disambiguate a target image from distractor images. A naive LM may generate the literal description  ``the red shape''; this is semantically accurate, but fails to disambiguate the referent among its context. A more pragmatically useful description is ``the red circle''.

Prior work in training communicative NLP systems has used explicit models of pragmatics \cite{goodman2016pragmatic} to fine-tune traditional LMs \cite{monroe2017colors,andreas2016reasoning, vedantam2017context}, or has trained LMs with external reward signals that indicate the contextual utility of an utterance. Human preferences are an ideal source of supervision for the latter approach \cite{stiennon2020learning, ziegler2020finetuning}, but are expensive to collect. 
One promising avenue is communication-based training (or \emph{self play}), where a speaker learns to communicate with a \emph{learned model} of a listener \cite{lazaridou2017multi,white2020learning}.
However, this approach commonly suffers from the problem of \emph{semantic drift}. Our use of ``semantic drift'' follows the communication-based training literature \cite{Lazaridou2020, lee2019countering}: the speaker produces utterances that satisfy the listener but diverge from the semantics and conventions of natural language. 

It can be difficult to diagnose the nature of semantic drift in complex, open domain models.
In this paper, we isolate one form of semantic drift that occurs when scaling communication-based training to larger domains: by increasing the size of the vocabulary available to the speaker and listener, the speaker fails to generalize to new listeners.
We identify miscalibrated listener uncertainty as the source of this problem:
the listener is highly confident in the interpretation of utterances outside its training domain, so the speaker overfits and produces nonsensical utterances that fail to generalize.
We propose to correct the calibration problem by using \emph{populations} of listeners, drawing from both computational and cognitive science work that suggests deficiencies in just training with a single listener \cite{graesser2020emergent, raviv2019larger,wagner2003progress} and improvements in emergent communication protocols by regularization effects of ensembled models  \cite{li2019ease,tieleman2019shaping}.
We find that ensemble-based populations of listeners are better calibrated, helping speakers avoid semantic drift and generalize to new games and listeners.

\section{Approach \label{sec:setup}}
We study the problem of learning a pragmatic speaker for reference games with the ShapeWorld \cite{kuhnle2017shapeworld} dataset. A reference game $(\mathbf{I}, t)$ consists of $n$ images $\mathbf{I}=(i_{1}, \dots, i_n)$ and a target image $i_t$, with the index $t$ known only to the speaker. 
The speaker $S$ must produce an utterance $u$ which allows the listener $L$ to identify the target $t$ given the images.

Formally, given $u$, the listener $L$ is a distribution over possible targets in a reference game:
\begin{equation}
    L(t \mid u) \propto \exp(f_{\theta}(i_t)^{\top}g_{\theta}(u))
    \label{eq:l0}
\end{equation}
where $f_{\theta}$ and $g_{\phi}$ are the listener's image and language encoders, respectively. 

The speaker is then trained to produce an utterance for the listener given a game and desired target.
Specifically, $S$ is parameterized by a gated recurrent neural network (GRU) \cite{cho2014learning}:
\begin{equation}
    S(u \mid \mathbf{I}) = p_{\text{GRU}}(u|f_{\eta}(i_{t}), f_\eta(i_1) \dots, f_{\eta}(i_{n-1})).
        \label{eq:s1}
\end{equation}
where $f_{\eta}$ is the speaker's image encoder used to initialize the GRU hidden state. The speaker is trained to maximize the listener's probability of selecting the correct target given its utterance. Formally, the speaker's loss over a game $G$ is
\begin{equation}
        \mathcal{L}(S; L, \mathbf{I}, t) = -\log L(t \mid \hat{u}),\, \hat{u} \sim S(u \mid \mathbf{I}).
        \label{eq:s1_loss}
\end{equation}

This \textit{informative communication} objective has been used several different ways in the past. 
Rational Speech Act models of pragmatic language use \cite{goodman2016pragmatic} adopt it as a definition of speaker behavior, rather than an objective for training.
Studies of emergent multi-agent communication \cite{lazaridou2017multi} use this objective to jointly train the speaker and listener, or alternatively train a pragmatic speaker against a fixed listener \cite{white2020learning,Lazaridou2020}. 
We adopt the setting of \citet{white2020learning}:
we first train listener models on separate splits of ShapeWorld reference games, then train pragmatic speakers with the fixed listener models as the communication objective. The listeners and speakers do not share data splits among each other. \textbf{Training listener(s)} are used as the internal listener model for the speaker in Equation~\ref{eq:s1_loss}, and other \textbf{validation listeners} are held-out for speaker evaluation. All the speakers and listeners are trained on 15000 randomly generated reference games with a single target and 2 distractor images.
More details on the model architecture, training, data, and the supplementary code can be found in Appendices \ref{sec:code} and \ref{sec:training}.

\begin{table*}[t]
  \caption{Average speaker accuracy $\pm$ standard error (middle 4 columns) and speaker-generated tokens overlapping with ShapeWorld tokens at test time (left-most column, ``token overlap'') in \% averaged over 10 seeds. Accuracy is reported with either the training or validation listener (train $L$, val $L$), and on either training or validation set (train $\mathcal{D}$, val $\mathcal{D}$). Gray cells indicate the speaker unable to generalize to new listeners and emitting novel tokens.
  \label{table:overfitting}}
%   \footnotesize
  \centering
  \begin{tabular}{llcccccc}
    \toprule
                     & token space & train $L$-train $\mathcal{D}$    & train $L$-val $\mathcal{D}$  & val $L$-train $\mathcal{D}$  & val $L$-val $\mathcal{D}$  & token overlap\\
                    %  &  & ($\downarrow$ better)       & ($\uparrow$ better)   & ($\downarrow$ better) \\
    \midrule
    \multirow{5}{*}{} & small  & 99 $\pm$ 1.40   & 90 $\pm$ 1.40 & 80 $\pm$ 1.40 & 80 $\pm$ 1.40  & 100 $\pm$ 0.0 \\
    & large             & 98 $\pm$ 0.00   & 94 $\pm$ 0.01 & \graycell 37 $\pm$ 0.06 & \graycell 36 $\pm$ 0.06 & \graycell 1 $\pm$ 0.6\\
    \bottomrule
  \end{tabular}
\end{table*}

\section{The problem of semantic drift \label{sec:theproblem}}
\citet{white2020learning} found successful pragmatic language production from a model trained to communicate informatively in the ShapeWorld domain.
We first show that this success hinges on the restricted domain of utterances used.
We consider two settings where the speaker learns to select utterances for two listeners: both are trained on the ground-truth ShapeWorld data, but one only has access to the ShapeWorld vocabulary (15 tokens), and one has access to the entire GPT-2 vocabulary (51k tokens; \citealt{radford2019language}). Given either listener, we train the speaker using Equation~\ref{eq:s1}. We chose GPT-2 vocabulary, a byte-pair encoding (BPE) \cite{sennrich-etal-2016-neural}, to access a large tokenization space as we hope to extend communication-based training to using GPT-2 as the generation model. Since BPE is the typical choice for most large pretrained language models, it is representative of large vocabulary spaces and introduces the challenges we might encounter in communication-based training for large speakers. 

In Table~\ref{table:overfitting}, we report speaker communication accuracy with either the listener used for speaker training 
(train $L$) or a separately trained held-out listener (val $L$), and either reference games seen during training (train $\mathcal{D}$) or held-out games (val $\mathcal{D}$).
To measure the extent the speaker uses domain-relevant vocabulary, we additionally report the percentage of speaker-generated tokens that overlap with ShapeWorld captions, such as ``blue circle'' or ``red square''.

The results reveal two noticeable discrepancies between the small and large vocabulary settings. 
First, speakers trained over the large vocabulary space can successfully communicate with the training listener on new reference games -- that is, they generalize to new contexts.
However, they are unable to generalize to held-out listeners and achieve only random chance accuracy ($33\%$).
This pattern of overfitting to the training listener despite generalizing to new contexts with the training listener indicates that the speaker suffers a form of \textit{semantic drift}  \cite{Lazaridou2020, lee2019countering}.

This semantic drift is shown most clearly in the second discrepancy:
speakers trained with large vocabularies rarely use tokens related to shapes and colors (i.e. the domain-relevant tokens), whereas those trained with small vocabularies do so (by design). Upon qualitative examination of the speaker's utterances (Table~\ref{tab:utterance_examples}), we observe some examples of this drift: the speaker's language deviates from natural language and the ShapeWorld domain that the neural listeners were trained on. 

\section{Diagnosing semantic drift \label{sec:diagnosing}}
We hypothesize that the observed semantic drift arises from poor uncertainty calibration in the neural listener, which the speaker then overfits to.
If this is correct, then populations of listeners---which have been shown to be better calibrated \cite{beluch2018power,li2019ease,tieleman2019shaping}---may yield more useful listeners for communication-based training.
To investigate this hypothesis we first explore the calibration of both single listeners and populations of listeners.

We compare the uncertainty measurements of three different internal listener implementations.
One baseline is \textbf{single-}$\mathbf{L_0}$, which approximates the internal listener $L$ with a single neural listener, i.e.\ exactly the setup in Section~\ref{sec:theproblem} where existing work uses a single neural listener for communication-based training. 
The second is an \textbf{ensemble-based population} of $n$ neural listeners; this is equivalent to substituting $L(t \mid u)= \frac{1}{n}\sum_{j=1}^n L^j(t \mid u)$) in Equation~\ref{eq:s1_loss} where each listener $L^j$ is initialized randomly. They are trained on different data splits and different random seeds, and each achieves at least 92\% accuracy on training and validation. While all listeners generalize to new  in-distribution utterances, they generalize differently out-of-distribution. It is these disagreements among the ensemble that improve calibration compared to a single neural network like  single-$L_0$.
The third is the \textbf{dropout-based population}, where we use a single listener to approximate a population of $n$ listeners via MC-dropout \cite{gal2016dropout}; this is equivalent to substituting $L(t|u)= \frac{1}{n}\sum_{j=1}^n L^{\texttt{dropout}}(t \mid u, d_j)$ in Equation~\ref{eq:s1_loss} where $d_j$ is a randomly sampled dropout mask. In our experiments, we use a dropout rate of $p=0.1$ but found no differences with using larger rates while maintaining high listener accuracy; please refer to Appendix~\ref{sec:training} for more details on the dropout implementation.

\begin{figure}[t]
    \centering
    \newcommand{\gw}{70mm}
    \newcommand{\plotr}{0.90}
    \begin{subfigure}[b]{\plotr\columnwidth}
      \centering
        \includegraphics[width=\gw]{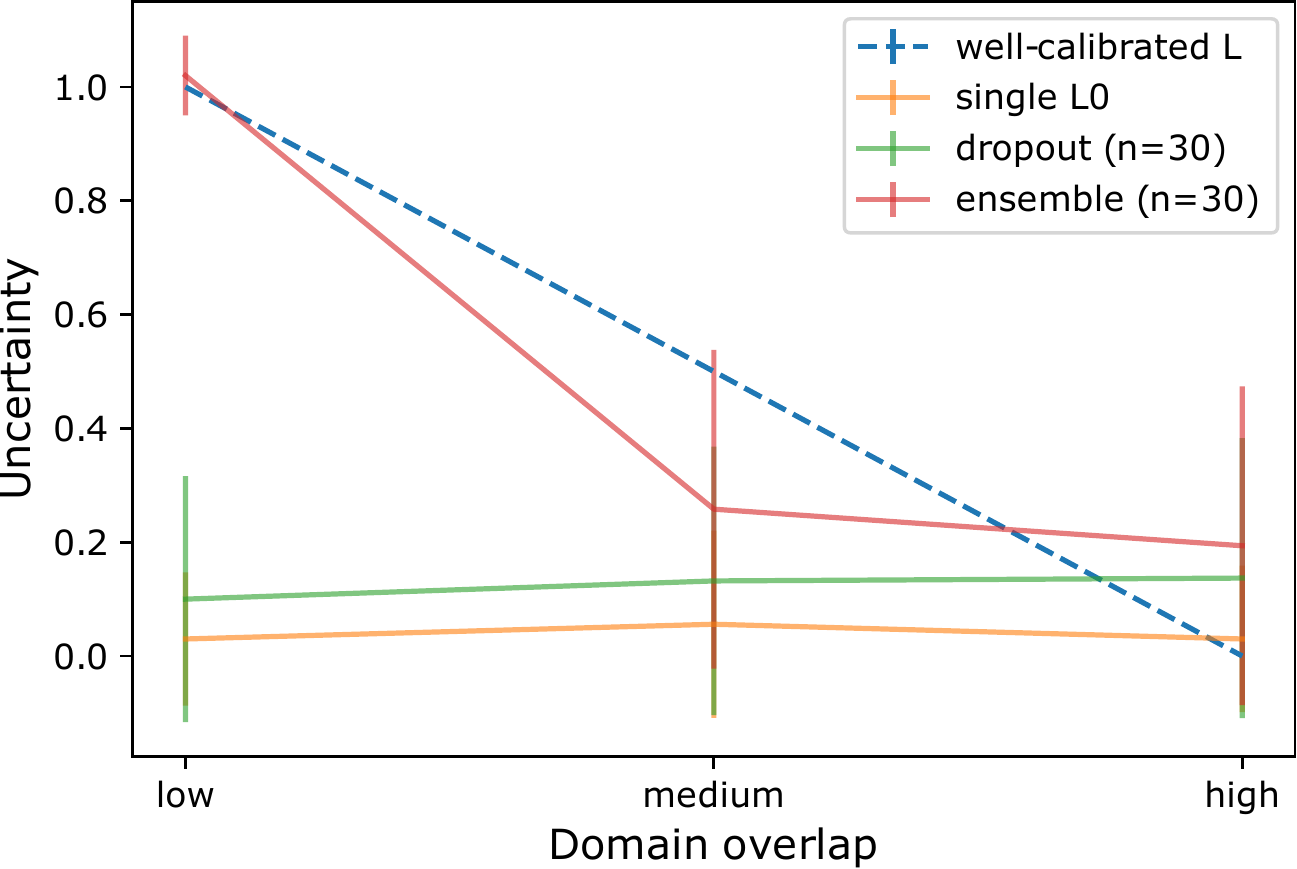}
    \end{subfigure}
    % \vspace{-1em}
    \caption{Listener uncertainty (entropy) by domain overlap. 
    Scores come from either single $L_0$, dropout-based population, or an ensemble-based population. Dashed line is an idealized listener, for reference.
    \label{fig:ood}}
\end{figure}

\begin{figure*}[h!]
    \centering
    \newcommand{\gw}{80mm}
    \newcommand{\plotr}{0.98}
    % \vspace{-1em}
    \begin{subfigure}[b]{\plotr\columnwidth}
        \centering
        \includegraphics[width=\gw]{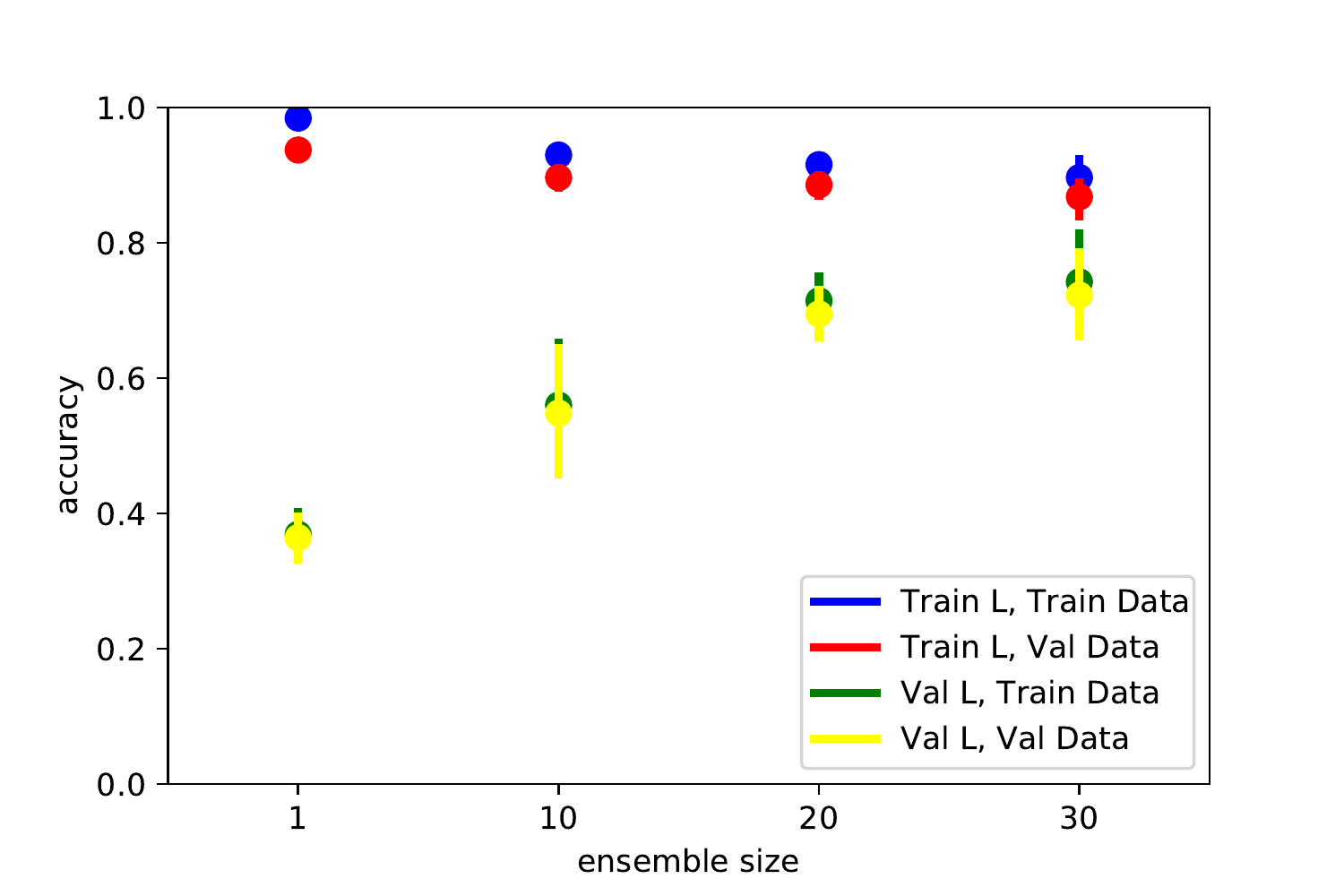}
        \caption{\label{fig:ensemble} Ensemble-based population training}
    \end{subfigure}
    \begin{subfigure}[b]{\plotr\columnwidth}
        \centering
        \includegraphics[width=\gw]{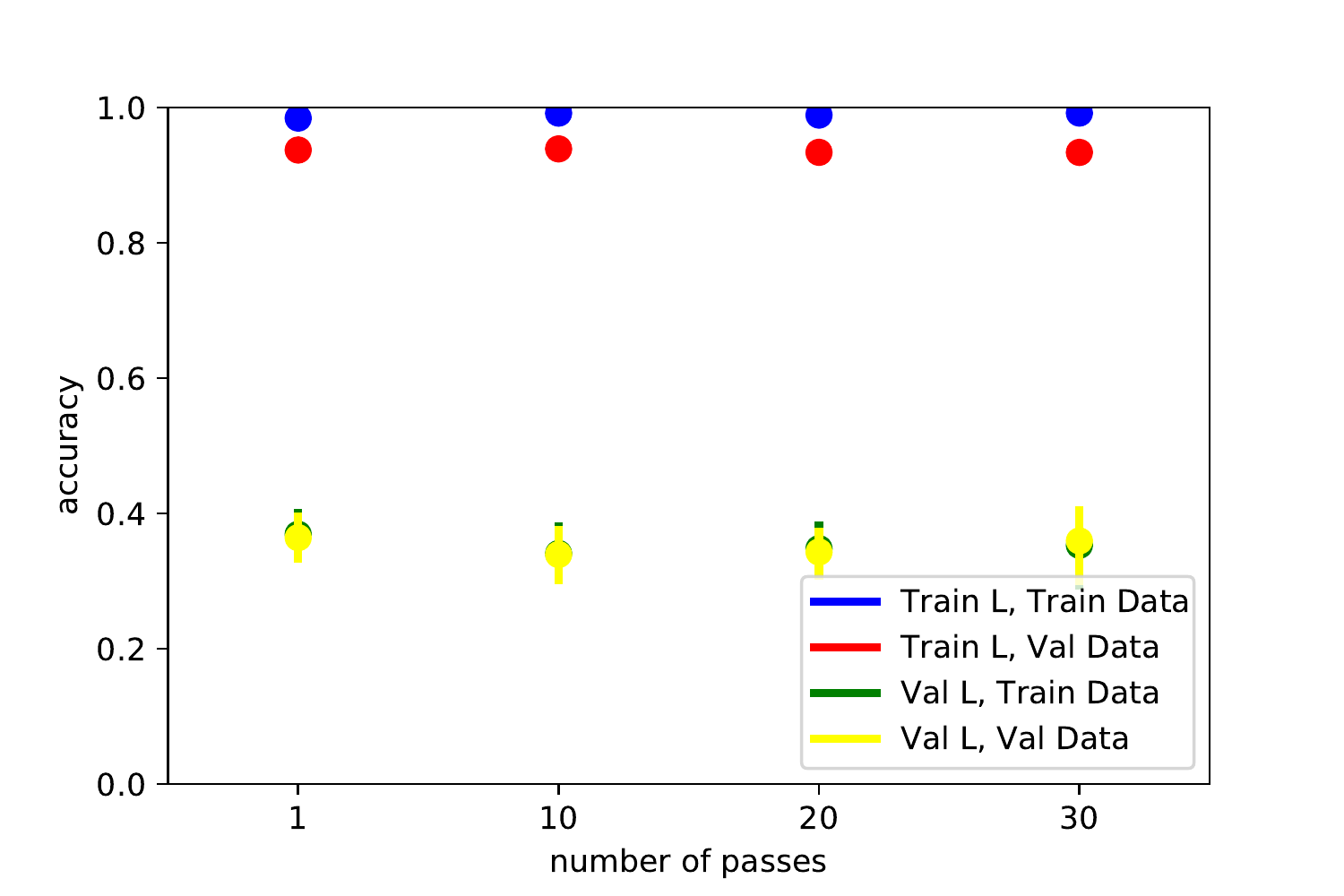}
        \caption{\label{fig:dropout} Dropout-based population training}
    \end{subfigure}
    \caption{Accuracy and standard error for speaker trained with  a (a) listener ensemble of size $n=\{1, 10, 20, 30\}$ or (b) dropout listener with $n=\{1, 10, 20, 30\}$ number of passes. $n$ represents the population size. Results averaged over 10 seeds.
    \label{fig:ensemble_sw}}
\end{figure*}

\begin{table}[h!]
%\hspace{-0.4cm}
    \footnotesize
    \begin{tabular}{p{1.9cm} l}
    \toprule
    \centering
    Target &
    \hspace{1cm} Distractors \\
   \hspace{0.1cm}  \includegraphics[scale=0.2]{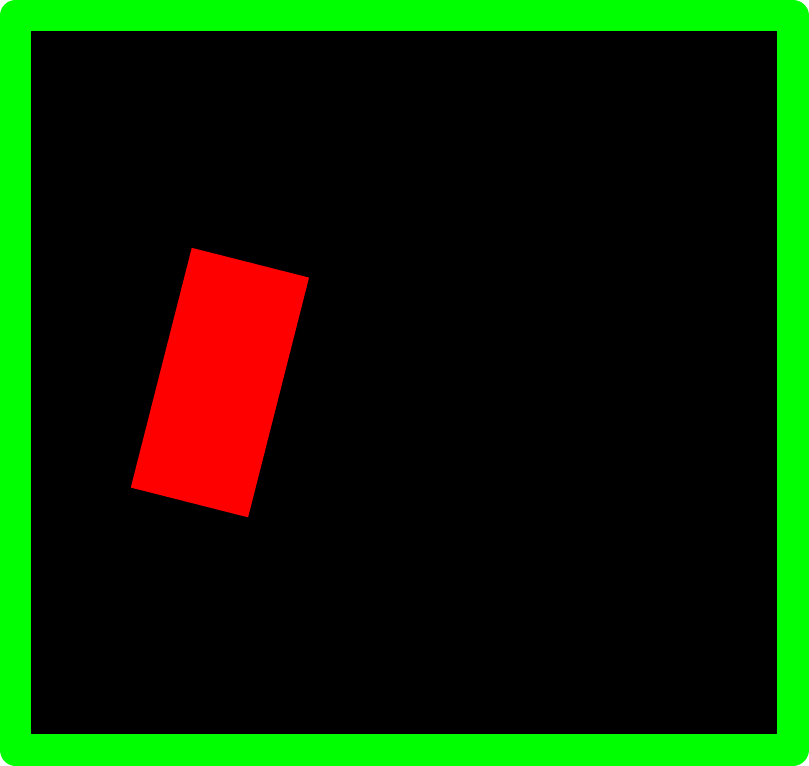} & 
    \hspace{1cm}\includegraphics[scale=0.20]{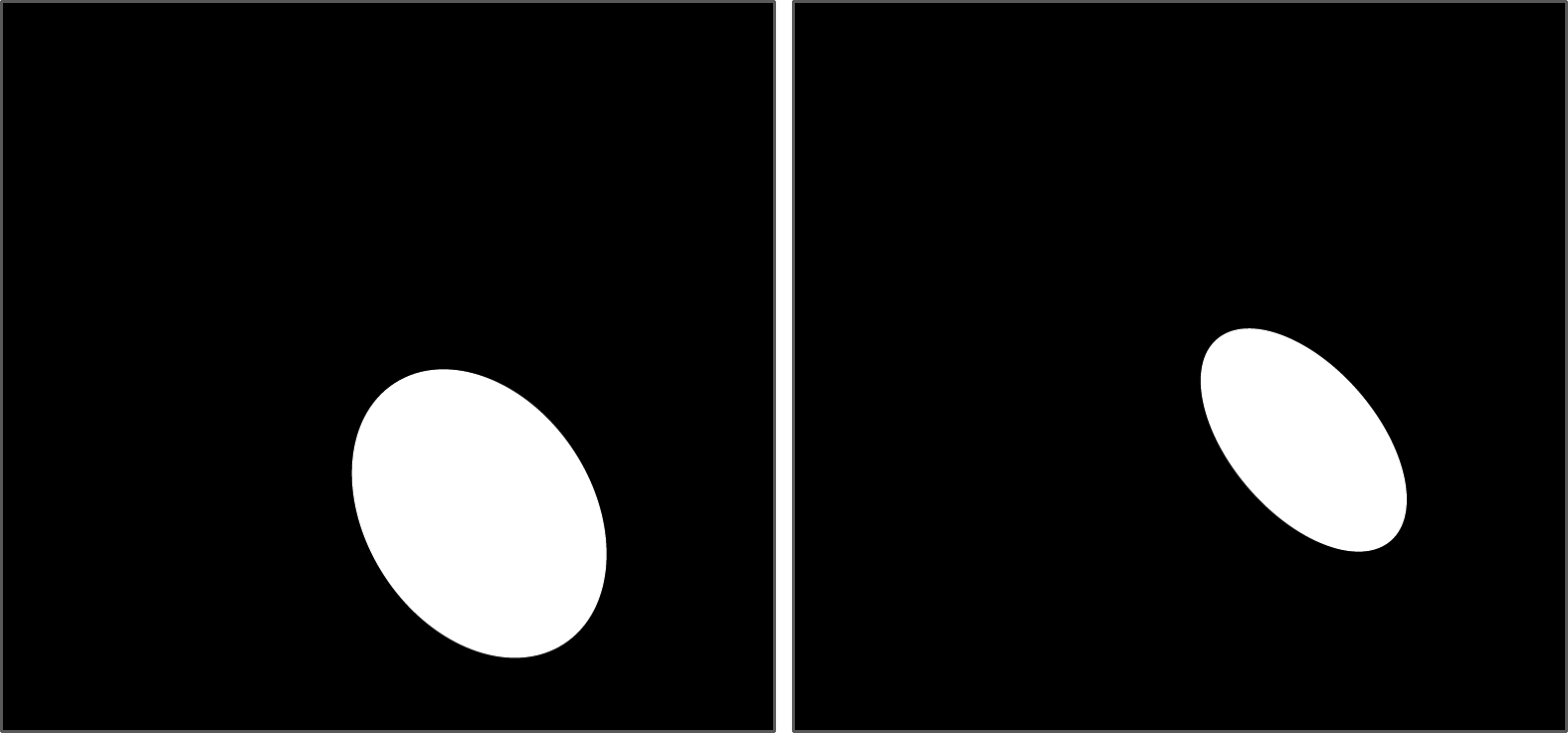}
    \\\midrule
    \multicolumn{2}{l}{\bf Method} \\
    % ~~\textbf{Speaker} & \textbf{Utterance} \\
    ~~gt  &  red shape \\
    ~~single $L_0$ & suppressed suppressed Mix \\ &  suppressed Pagan Mix Mix\\ &  Pagan Mix \\
    ~~dropout & suppressed suppressed \\ &  suppressedops suppressed Mix \\ &  Pagan imprison Mix \\
    ~~ensemble & rect rect rect rect \\[-0.5em] \\ \cline{1-2} \\[-0.5em]
    \centering
    Target &
    \hspace{1cm} Distractors\\
     \hspace{0.1cm} \includegraphics[scale=0.2]{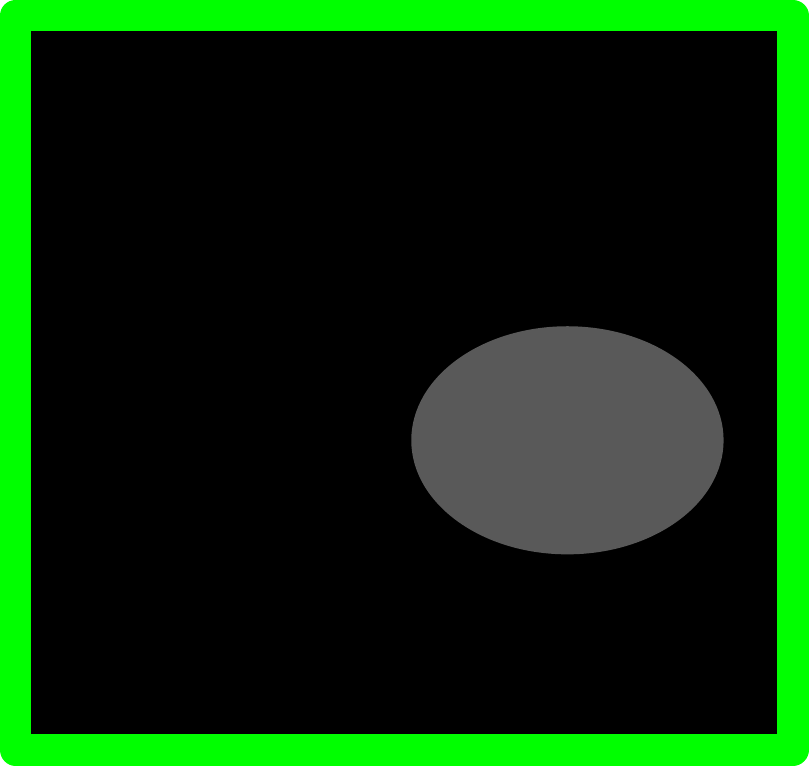} & 
    \hspace{1cm}\includegraphics[scale=0.20]{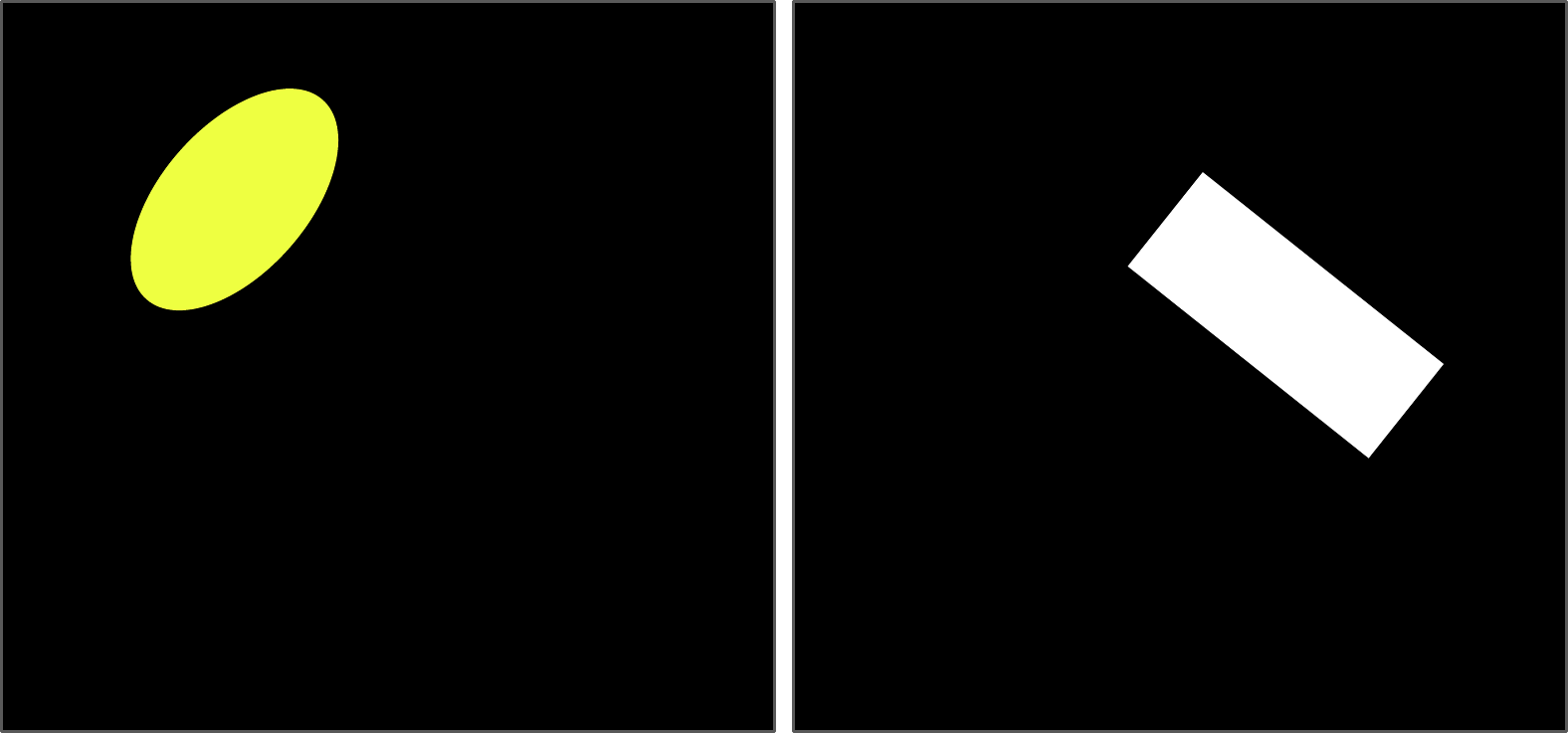}
    \\\midrule
    \multicolumn{2}{l}{\bf  Method}  \\
    ~~gt  &  gray shape \\
    ~~single $L_0$ & ESEESE FDAWHAT \\&Macedichickarios \\&Macedichick \\
    ~~dropout & ariosichickarios FDA FDAarios \\ &brainsichick Maced \\
    ~~ensemble & gray gray gray \\[-0.7em] \\\cline{1-2} \\[-0.7em] \\
    \end{tabular}
    % \vspace{-1em}
\caption{Examples of generated utterances. ``gt'' refers to the ground truth ShapeWorld utterance associated to the target image (left-most image). We show the utterance generated by the speaker trained with either the single $L_0$, dropout-based population, or ensemble-based population method. Based on token overlap and the reference game context, the ensemble-based population method results in more pragmatic and useful utterances.}
\label{tab:utterance_examples}
    
\end{table}

Intuitively, the ideal listener should be uncertain about utterances that are different from their training domain.
To measure this, we evaluate listener uncertainty scores on utterances with varying degrees of overlap with their training domain. Specifically, for \textbf{high-overlap} utterances, we sample messages directly from the ground truth ShapeWorld language. For \textbf{medium-overlap} utterances, we sample language from a well-calibrated neural speaker (Table~\ref{tab:utterance_examples}, ensemble). For \textbf{low-overlap} utterances, we sample from a poorly-calibrated speaker (Table~\ref{tab:utterance_examples}, single $L_0$). Refer to Appendix~\ref{sec:utterance_domain} for a complete discussion on utterance sampling. For these utterance types, the ideal listener should have maximum entropy over targets for low-overlap utterances, and zero entropy for high-overlap utterances.
In Figure~\ref{fig:ood}, we plot the listener's uncertainty over utterances with varying domain overlap.
We see, as hypothesized, that the single listener is poorly calibrated, yielding very low entropy for low-overlap utterances.
Somewhat surprisingly, dropout-based populations do not help either.
The ensemble-based populations, however, do show better calibration and are closer to the idealized listener: they are uncertain about out-of-domain utterances.

\section{Countering semantic drift} Following our hypothesis, a well-calibrated listener should lead the speaker to generalize to new listeners \textit{and} new contexts while reasoning pragmatically over a large vocabulary. We show that this is indeed the case in the following two experiments. 

First, we show that with increasing population size $n$, the ensemble-based population objective closes the accuracy gap between training and validation listeners, while still generalizing to new reference games (Figure~\ref{fig:ensemble_sw}). This is not the case for the dropout listener with increasing the number of passes $n$. These results are averaged on 10 seeds, for $n$ of $\{1,10,20,30\}$. With ensemble-based population $n=30$, we found that the speaker token overlap increased to $48.4 \pm 12.3 \%$; this is a much larger token overlap than the original speaker trained on single-$L0$ ($1.0 \pm 0.6$ from Table~\ref{table:overfitting}). We show utterance examples from speakers trained with the population objectives in Table~\ref{tab:utterance_examples}.

\begin{table}[h]
  \footnotesize
  \centering
  \begin{tabular}{llccc}
    \toprule
                     & speaker type & $d(z_{\text{GT}}$, \texttt{SUM}$(z_{\text{S}}))$ & $d(z_{\text{GT}}$, \texttt{FIRST}$(z_{\text{S}}))$\\
                    %  &  & ($\downarrow$ better)       & ($\uparrow$ better)   & ($\downarrow$ better) \\
    \midrule
    \multirow{3}{*}{} & Limited       & 9.2 $\pm$ 3.74               & 6.43 $\pm$ 2.54               \\ 
    & Calibrated             & 10.37 $\pm$ 2.41              & 8.52 $\pm$ 0.50  \\
    & Miscalibrated             &  13.03 $\pm$ 3.62             & 8.72 $\pm$ 0.44  \\
    \bottomrule
  \end{tabular}
  \caption{ Euclidean distance $d$ in GloVe embedding space between the ground truth utterances $z_{\text{GT}}$ and speaker utterances $z_{\text{S}}$. Lower distances indicate that the speaker's utterance is close in topicality to the ground truth. The speakers are either trained only on the Shapeworld vocabulary (Limited), trained with an ensemble-based listener population (Calibrated), or trained with single-$L0$ (Miscalibrated). The speaker utterance embeddings are either calculated as a sum of the uttered words (\texttt{SUM}$(z_{\text{S}})$) or as the embedding of the first uttered word (\texttt{FIRST}$(z_{\text{S}})$). 
  \label{table:topic}}
\end{table}

To explore the semantic content of utterance distributions, we investigate the topic overlap between different speakers and the ground truth utterances.
We find that speakers trained with well-calibrated listeners produce messages which are semantically similar to the ground truth utterances, according to distances in GloVe \cite{pennington2014glove} embedding space (Table~\ref{table:topic}). We embed the ground truth utterances by taking the sum of embeddings over the ground truth utterance. We denote this sum as $z_{\text{GT}}$. We compare these to the utterance embeddings of three speakers: one speaker which is constrained to Shapeworld vocabulary (used in \citet{white2020learning} and in Section~\ref{sec:theproblem}), one that is trained with well-calibrated listeners (ensemble-based population of size 30), and another that is trained with the poorly calibrated single-$L0$ listener. The speaker's embeddings, $z_{{\text{S}}}$, are calculated either as a sum of uttered word embeddings or as the embedding of the first utterance word; we include the results from the latter because  miscalibrated speakers often repeat words and taking only the sum might distort the calculated embedding distance. The distance reported in Table~\ref{table:topic} is the Euclidean distance between $z_{\text{GT}}$ and $z_{{\text{S}}}$. The results show that the calibrated speaker utters words that are similar to ground truth utterances, though not quite as similar as the limited speaker. The miscalibrated speaker utters words that are less similar to the ground truth utterances as indicated by the higher distance scores.

%%%%%%%%%%%%%%%%%%%%%%%%%%%%%%%%%%%%
\section{Conclusion}
The true objective for language use is communication. 
Our work highlights the importance of well-calibrated listeners in communication-based training. 
We show that it's important to understand the properties of speakers optimized to communicate with different listeners.
We have found that naive communication-based training over unconstrained vocabularies is subject to a pernicious form of semantic drift that arises from poorly calibrated listener models. The overconfidence of listeners is exploited by the speaker during training and, as a result,  speakers acquire niche linguistic properties, like conventions that fail to generalize to other listeners \cite{hawkins2017convention, graesser2020emergent}.  
By contrast, an ensemble of listeners shows better calibration. Speakers trained to communicate with these listeners avoid semantic drift, generalizing to new games and new listeners.
Future research on communication-based training for language models will thus benefit from listener ensembles or other methods for training listeners with properly calibrated uncertainty.

% \newpage
 
\section{Acknowledgements}

REW and JM are supported by the National Science Foundation Graduate Research Fellowship. This work was supported in part by the Stanford HAI Hoffman--Yee project `Towards grounded, adaptive communication agents.'
We thank CoCoLab and the anonymous reviewers for their helpful feedback on the paper, and Ke Alexander Wang for helpful input on Figure 1. 

\bibliography{anthology,paper}
\bibliographystyle{acl_natbib}

\clearpage
\newpage
% \mbox{~}
\clearpage
\newpage

\appendix

\section{Software and data for reproducibility \label{sec:code}}
The accompanying code can be found here: \url{https://github.com/rosewang2008/calibrate_your_listeners}. It includes code for training the models, instructions, the data used for this work, and pretrained $L$ models.

\section{Implementation and training \label{sec:training}}

\paragraph{Model details} 
The listener and speaker have a separate vision encoder, $f_{\theta}$ and $f_{\phi}$, which are convolutional neural networks with 4 blocks, each block containing a 64-filter 3x3 convolutional layer, batch
normalization, ReLU activation, and 2x2 max pooling. Images are represented as 64×64×3 input, which produce 1024-d representations. For speakers, the image embeddings are projected down to the GRU hidden state size for hidden state initialization in $p_{\text{GRU}}$. For listeners, we take the dot product of the image embeddings and utterance embeddings to produce the target probability. The listener's language encoder $g(\cdot)$ is also a gated RNN. The listener and speaker's GRU has two layers. It takes in an embedding dimension of 512 and has a hidden size of 100.  
The token space varies in our experiments: the small vocabulary setting has 15 tokens as in \citet{white2020learning} and the large setting has about 51k tokens from the GPT2 tokenizer \cite{wolf2020huggingfaces, radford2019language}. The speaker's maximum sequence length is 10 tokens.

For the dropout-based listener, we apply a dropout layer with dropout rate $p=0.1$ in the listener's image and language encoders, $f_{\theta}, g_{\theta}$. Dropout is applied after every convolutional block except the last block in the image encoder. Dropout is applied on the outputs of each GRU layer except for the last layer. We experimented with higher dropout rates ($p=0.2, 0.3$) while maintaining high listener validation accuracy during listener evaluation, and found no differences in the reported experimental findings. Specifically, we found that using listeners with higher dropout rates did not change the uncertainty measurements shown in Figure~\ref{fig:ood}. Speakers trained with listeners of higher dropout rates still obtained training and validation accuracies that were still within the margin of error reported in Figure~\ref{fig:ensemble_sw}.

The listeners in the ensemble-based population do not share any encoders and are trained on separate models.

\paragraph{Training} All the models were trained on 100 epochs, with the Adam optimizer, a batch size of 32, using the Gumbel Softmax trick \cite{Jang2017}. The learning rate is 0.001 for speakers and 0.01 for listeners. All the listeners after training reach at least 92\% validation accuracy, and are used for speaker training or validation.

\section{Domain overlap in utterances \label{sec:utterance_domain}}
Intuitively, we want utterances with low domain overlap to be utterances that are unrelated ShapeWorld domain (e.g. those sampled from the speaker trained with the single-$L_0$ method as shown in Table~\ref{tab:utterance_examples}), and utterances with high domain overlap to be captions correct for a given target image and perfectly related to the ShapeWorld domain and correct (e.g. those directly from the ShapeWorld dataset). Utterances with medium domain overlap should be an interpolation between low and high domain overlap. In our setting that is concerned with semantic language drift, we define overlap to be with respect to the percentage of utterance tokens that are ShapeWorld-related tokens.

We use the ShapeWorld image captions as the high domain overlap because these captions consist of only tokens related to the ShapeWorld setting, i.e. $100\%$ overlap. We use utterances sampled from the speaker trained with ensemble-based population for the medium overlap utterances; these utterances generally use ShapeWorld-related tokens but in repeated fashion (see Table~\ref{tab:utterance_examples} for an example) and have $51.6\%$ overlap. We use utterances sampled from the speaker trained with the single-$L_0$ method with $0\%$ token overlap for the low domain overlap, as we found that this condition applied to a wide majority of utterances sampled from this speaker. We note that using token overlap as a proxy for domain overlap is an imperfect measurement as it doesn't factor in whether an utterance is actually appropriate for a desired target image. For example, an utterance like ``red green shape'' uses ShapeWorld-related tokens, but may not be an ideal utterance for a red-shape target image. Nonetheless, since our work is concerned with semantic language drift that uses novel and out-of-domain tokens, we found this to be the best measurement of domain overlap compared to alternatives like n-gram overlap in tokens.

\end{document}